\documentclass{article}
\usepackage{ijcai26}

\usepackage{times}
\usepackage{soul}
\usepackage{url}
\usepackage[hidelinks]{hyperref}
\usepackage[utf8]{inputenc}
\usepackage[small]{caption}
\usepackage{graphicx}
\usepackage{amsmath}
\usepackage{amsthm}
\usepackage{booktabs}
\usepackage{algorithm}
\usepackage{algorithmic}
\usepackage[switch]{lineno}
\usepackage[table]{xcolor}
\usepackage{amssymb}

\newcommand{\first}[1]{\cellcolor{blue!30!white}{#1}}
\newcommand{\second}[1]{\cellcolor{blue!20!white}{#1}}
\newcommand{\third}[1]{\cellcolor{blue!10!white}{#1}}

\newcommand{\bfirst}[1]{\colorbox{blue!30}{#1}}
\newcommand{\bsecond}[1]{\colorbox{blue!20}{#1}}
\newcommand{\bthird}[1]{\colorbox{blue!10}{#1}}

\title{Physic-HM: Restoring Physical Generative Logic in Multimodal Anomaly Detection via Hierarchical Modulation}

\author{
Xiao Liu$^{1}$\and
Junchen Jin$^{2}$\and
Yanjie Zhao$^{2}$\and
Zhixuan Xing$^{3}$\\
\affiliations
$^{1,2,3,4}$Chongqing University\\
\emails
liu-xiao-@outlook.com
}

\begin{document}

\maketitle

\begin{abstract}

Multimodal Unsupervised Anomaly Detection (UAD) is critical for quality assurance in smart manufacturing, particularly in complex processes like robotic welding. However, existing methods often suffer from process-logic blindness, treating process modalities (e.g., real-time video, audio, and sensors) and result modalities (e.g., post-weld images) as symmetric feature sources, thereby ignoring the inherent unidirectional physical generative logic. Furthermore, the heterogeneity gap between high-dimensional visual data and low-dimensional sensor signals frequently leads to critical process context being drowned out. In this paper, we propose Physic-HM, a multimodal UAD framework that explicitly incorporates physical inductive bias to model the $Process \to Result$ dependency. Specifically, our framework incorporates two key innovations: a Sensor-Guided PHM Modulation mechanism that utilizes low-dimensional sensor signals as context to guide high-dimensional audio-visual feature extraction, and a Physic-Hierarchical architecture that enforces a unidirectional generative mapping to identify anomalies that violate physical consistency. Extensive experiments on Weld-4M benchmark demonstrate that Physic-HM achieves a SOTA I-AUROC of 90.7\%. The source code of Physic-HM will be released after the paper is accepted.
\end{abstract}

\section{Introduction} 

\begin{figure}[htb]
    \centering
    \includegraphics[width=\linewidth]{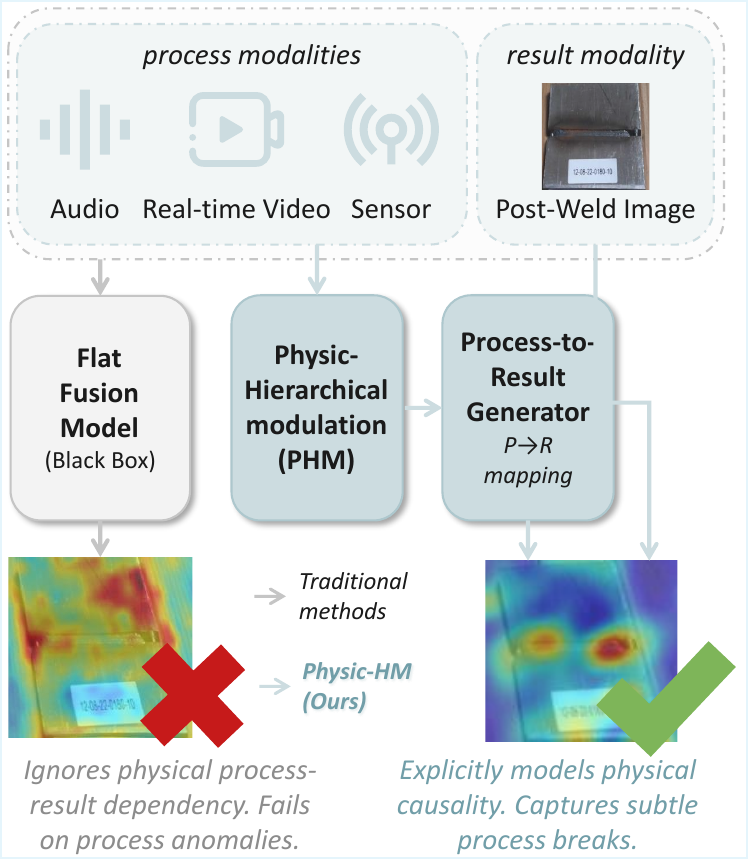}
    \caption{Comparison between traditional flat fusion and our physical-Hierarchical Fusion.}
    \label{fig:teaser}
    \vspace{-2mm}
\end{figure}

Industrial Anomaly Detection (IAD) has become a fundamental pillar of modern smart manufacturing, providing the necessary oversight for quality assurance in complex production environments \cite{liu2024deep,kim2024rethinking}. In high-precision domains such as robotic welding, identifying rare and diverse defects is not only a safety requirement but also a prerequisite for critical downstream tasks \cite{hong2024af,cheng2023intelligent}. These include real-time closed-loop control of autonomous production lines, structural integrity assessment for aerospace components, and the optimization of resource allocation in automated factories \cite{pemula2025robust}. Given the extreme scarcity of defect samples in real-world settings, Unsupervised Anomaly Detection (UAD) has emerged as the mainstream paradigm, aiming to learn the distribution of normal patterns and identify any deviation as a potential anomaly \cite{lin2025survey}.

To capture the multi-faceted nature of industrial processes, recent research has shifted from single-modal image analysis to multimodal fusion strategies \cite{zhang2024realnet}. Current state-of-the-art (SOTA) methods primarily follow two paths: feature-embedding-based \cite{zhang2024contextual} and reconstruction-based \cite{cheng2025mc3d} methods. For instance, M3DM \cite{wang2023multimodal} utilizes hybrid memory banks to align RGB and point-cloud features, while Dinomaly \cite{guo2025dinomaly} leverages high-resolution frozen backbones and transformer architectures to suppress the reconstruction of anomalies. Furthermore, several works have explored the integration of sensor emissions and real-time video to enhance detection robustness in noisy environments \cite{stemmer2024unsupervised,wu2024cross}. These advancements have significantly improved performance on standard visual benchmarks.

However, existing multimodal methods face a critical limitation we define as process-logic blindness. As illustrated in Figure~\ref{fig:teaser}, traditional flat fusion strategies treat all available modalities—such as video, audio, and sensor time-series—as symmetric feature sources \cite{you2022unified}. This symmetry fundamentally ignores the inherent physical generative logic of industrial production: the Process (e.g., arc current, welding sound) is the cause, and the Result (e.g., the final weld bead surface) is the effect. Although some methods attempt to balance these signals via information-theoretic feature integration \cite{gao2024embracing}, they still suffer from the heterogeneity gap, where critical but low-dimensional sensor semantics are drowned out by high-dimensional visual features \cite{wang2023multimodal,he2024mambaad}. For example, a lack of fusion defect might be caused by a momentary drop in current (identifiable in sensors) but result in a visually normal-looking surface. A flat fusion model, biased toward visual dominance, would likely ignore the sensor anomaly and misclassify the sample as normal.



To overcome these challenges, we propose Physic-HM, a multimodal UAD framework that incorporates a physical generative inductive bias to model the unidirectional $P \rightarrow R$ dependency. Departing from traditional symmetric fusion, our Physic-Hierarchical architecture explicitly treats process modalities (real-time video, audio, and sensors) as physical governors and the result modality (post-weld images) as the observable effect. We design a Sensor-Guided PHM Modulation module that utilizes Mamba-encoded sensor signals to scale and shift audio-visual feature extraction via affine transformations. This ensures that low-dimensional process constraints are injected into high-dimensional representations without being overwhelmed. An anti-generalization decoder then reconstructs the result latent from these modulated features, while a semantic text prior anchors the manifold of normality. By enforcing the physical laws of normal production, any violation in the generative mapping $f: P \rightarrow R$ triggers a detectable anomaly, even if the final product appears visually nominal. Experimental results on the Weld-4M benchmark demonstrate a SOTA I-AUROC of 90.7\%. Notably, Physic-HM achieves nearly 6x faster inference than memory-bank-based methods, proving its operational feasibility.
In summary, the contributions of this paper are:
 \begin{itemize}
	\item We introduce a Physic-Hierarchical architecture that replaces symmetric feature fusion with a directional generative prior, enabling the detection of deep-seated process anomalies that are invisible to purely visual models.
	\item We propose the PHM modulation mechanism, which leverages Mamba-based temporal encoders to transform low-dimensional sensor signals into contextual priors. This design scales and shifts audio-visual features via affine transformations, preventing critical process context from being drowned out by high-dimensional visual data.
	\item Extensive evaluations on the Weld-4M benchmark demonstrate that Physic-HM achieves a SOTA I-AUROC of 90.7\%. Our approach shows superior sensitivity and robustness to process-hidden defects and maintains nearly 6x faster inference than traditional memory-bank-based methods.
\end{itemize}

\section{Related Work}

\begin{figure*}[t]
\centering
\includegraphics[width=1\textwidth]{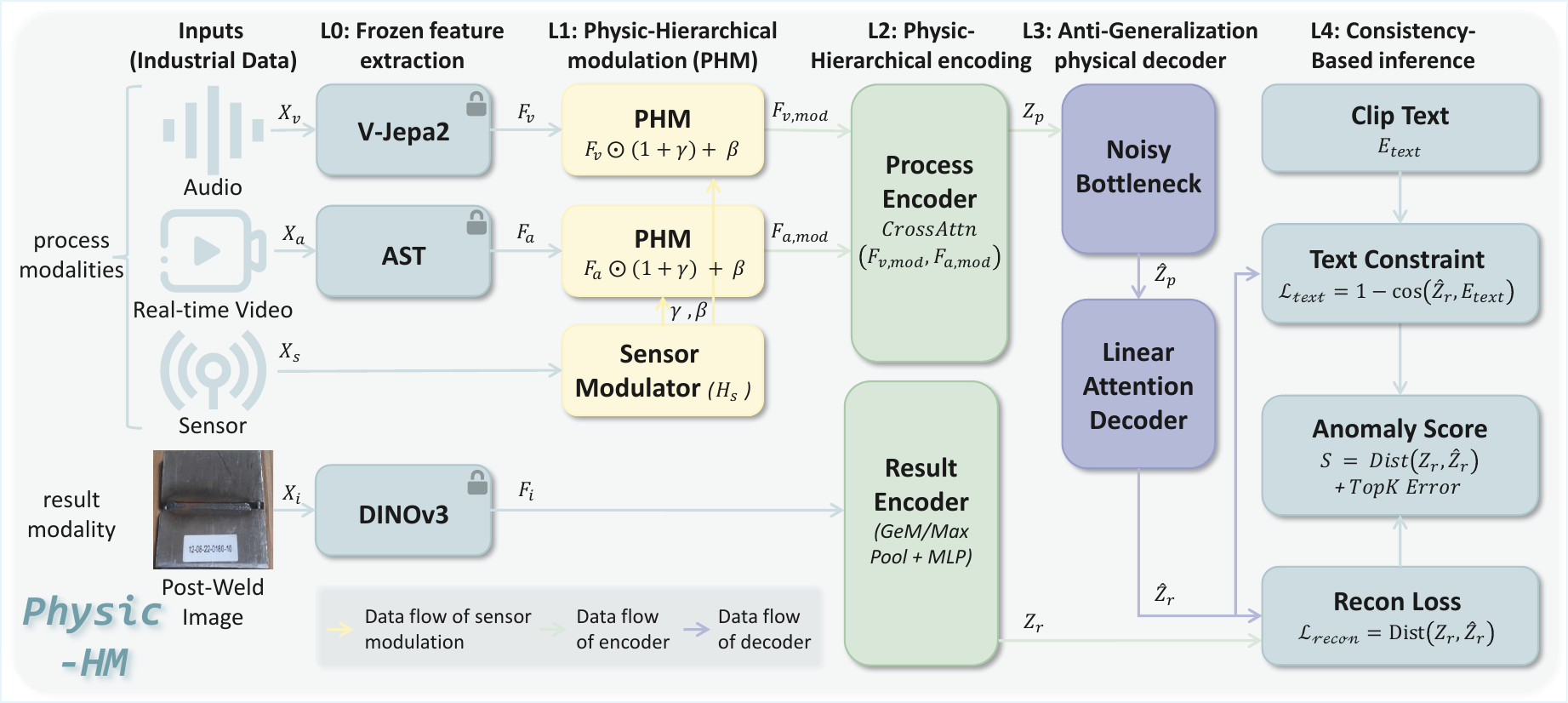}
\caption{The overall architecture of Physic-HM. The framework operates on fixed representations extracted by frozen backbones (L0, in blue) and is organized into a hierarchical physical logic structure. (a) Functional Pathways: distinct colors denote trainable components, including physical encoders (green), sensor-guided PHM modulation (gold), and the anti-generalization decoder (purple). (b) Mechanism: The model explicitly learns the unidirectional physical mapping $P \to R$ to detect anomalies that violate production consistency.}
\label{fig:architecture}
\vspace{-2mm}
\end{figure*}

\subsection{Reconstruction-based Anomaly Detection}
Industrial Anomaly Detection (IAD) has witnessed a rapid transition from single-modal visual inspection to unified multimodal paradigms to meet the reliability demands of smart manufacturing \cite{liu2024deep}. Current state-of-the-art (SOTA) reconstruction methods, such as Realnet \cite{zhang2024realnet} and OCR-GAN \cite{liang2023omni}, have leveraged adaptive feature selection network and channel selection to suppress the reconstruction of anomalous patterns through anti-generalization mechanisms \cite{guo2023recontrast}. To handle more complex industrial scenarios, multimodal frameworks like M3DM \cite{wang2023multimodal} and SiM3D \cite{costanzino2025sim3d} have successfully integrated RGB-D or point-cloud data using hybrid memory banks or cross-modal contrastive learning. Furthermore, the emergence of generalist anomaly detection—exemplified by AnomalyCLIP \cite{zhou2023anomalyclip}, UNPrompt \cite{niu2024zero}, and in-context residual learning works \cite{zhu2024toward}, aims to achieve zero-shot generalization across diverse industrial domains \cite{liu2024arc}. Despite these advancements, existing multimodal methods predominantly adopt a symmetric fusion strategy, treating all signals (e.g., video, audio, and sensors) as parallel feature sources. This approach ignores the unidirectional physical generative logic where the industrial process dictates the resulting product, leading to process-logic blindness when process-stage anomalies do not leave immediate surface-level evidence. Our Physic-HM addresses this by explicitly incorporating physical inductive bias through a unidirectional $Process \rightarrow Result$ dependency, allowing the model to detect subtle physical consistency breaks that symmetric fusion models overlook.

\subsection{Multimodal Signal Modulation}
Efficiently integrating heterogeneous industrial signals remains a significant challenge due to the extreme information density imbalance between high-dimensional visual streams and low-dimensional, high-frequency sensor data \cite{wang2023multimodal,cheng2023intelligent,li2025multi}. Conventional fusion techniques, such as early concatenation, score-level late fusion \cite{stemmer2024unsupervised} or attention \cite{you2022unified}, often suffer from the modality drowning effect, where critical but low-dimensional process context is overwhelmed by visual features \cite{barua2023systematic}. While recent studies have explored pretrained CLIP, Point-BIND and Reverse Distillation \cite{gu2024rethinking} to mitigate modal noise \cite{wang2025m3dm}, they still treat sensor signals as auxiliary features rather than governing physical priors. Simultaneously, State Space Models (SSMs), particularly Mamba \cite{gu2024mamba}, have demonstrated superior efficiency in modeling long-range temporal dependencies with linear complexity, leading to breakthroughs in multi-class and online anomaly detection \cite{he2024mambaad}. However, current IAD frameworks have not fully exploited the potential of sensors to modulate higher-dimensional representations. We bridge this gap by proposing a Sensor-Guided PHM (Physic-Hierarchical Modulation) module built upon a Mamba temporal encoder. Unlike previous methods that perform flat feature integration, our approach utilizes sensor-driven affine transformations to dynamically scale and shift audio-visual features, ensuring the physical process state serves as a governor for the feature extraction of all modalities, preventing process-critical signals from being drowned out.

\section{Method}

\subsection{Preliminary}
The objective of unsupervised multimodal anomaly detection in a manufacturing context is to construct a model that learns the manifold of normal production dynamics from a defect-free training set. We denote the training dataset as $\mathcal{D} = \{X^{(n)}\}_{n=1}^N$, where each sample $X$ is a synchronized multimodal collection $X = \{X_v, X_a, X_s, X_i\}$.

Here, $X_v \in \mathbb{R}^{T_v \times 3 \times H \times W}$ represents the real-time process video capturing the spatiotemporal evolution of the work area, $X_a \in \mathbb{R}^{F \times T_a}$ denotes the acoustic emission spectrogram, and $X_s \in \mathbb{R}^{T_s \times C_s}$ refers to 1D sensor time-series such as current and voltage. These process signals are complemented by $X_i \in \mathbb{R}^{M \times 3 \times H \times W}$, representing the set of $M$ post-weld images capturing the final result from multiple camera perspectives.

Our fundamental hypothesis is that industrial production follows a unidirectional physical generative chain. We define the process-stage modalities as the Cause (Process) $P = \{X_v, X_a, X_s\}$ and the final outcome as the Effect (Result) $R = \{X_i\}$. An anomaly is thus defined not merely as a statistical outlier, but as a violation of the physical generative mapping $f: P \to R$, where the observed result $R$ becomes inconsistent with the result predicted from the observed process $P$.

\subsection{Overall Framework}

\begin{algorithm}[tb]
\caption{Physic-HM Pipeline}
\label{alg:physic_hm}
\begin{algorithmic}[1] 
    \renewcommand{\algorithmicrequire}{\textbf{Input:}}
    \renewcommand{\algorithmicensure}{\textbf{Output:}}
    
    \REQUIRE Process Video $X_v$; Acoustic Spectrogram $X_a$; Sensor Series $X_s$; Result Images $X_i$
    \ENSURE Anomaly Score $S$
    
    \STATE // \textsf{Frozen feature extraction (L0)}
    \STATE $F_v \leftarrow \text{V-JEPA2}(X_v), \ F_a \leftarrow \text{AST}(X_a), \ F_i \leftarrow \text{DINOV3}(X_i)$
    
    \STATE // \textsf{Physic-Hierarchical modulation (L1)}
    \STATE $h_s \leftarrow \text{Mamba\_SSM}(X_s)$
    \STATE $\gamma, \beta \leftarrow \text{Linear\_Projection}(h_s) \in \mathbb{R}^D$
    \STATE $F_{v,mod} = F_v \odot (1 + \gamma) + \beta$
    \STATE $F_{a,mod} = F_a \odot (1 + \gamma) + \beta$
    
    \STATE // \textsf{Physic-Hierarchical encoding (L2)}
    \STATE $Z_{p} \leftarrow \text{CrossAttention}(F_{v,mod}, F_{a,mod})$
    \STATE $Z_r \leftarrow \text{MLP}(\text{GatedAngleAggregation}(F_i))$
    
    \STATE // \textsf{Anti-Generalization Physical Decoder (L3)}
    \STATE $\hat{Z}_{p} \leftarrow \text{NoisyBottleneck}(Z_{p})$
    \STATE $\hat{Z}_{r} \leftarrow \text{LinearAttentionDecoder}(\hat{Z}_{p})$
    \STATE $\mathcal{L} = \mathcal{L}_{recon}(Z_r, \hat{Z}_r) + \lambda \mathcal{L}_{text}(\hat{Z}_r, E_{text})$
    
    \STATE // \textsf{Consistency-Based inference (L4)}
    \STATE $S = \text{CosineDist}(Z_{r}, \hat{Z}_{r}) + \eta \cdot \text{mean}(\text{TopK}(|Z_{r} - \hat{Z}_{r}|))$
    \STATE \textbf{return} $S$
\end{algorithmic}
\end{algorithm}
\vspace{-2mm}


The implementation of Physic-HM follows a hierarchical four-level architecture (L1–L4) designed to bridge the heterogeneity gap while preserving the unidirectional physical flow from the production process to the final result, as illustrated in Figure~\ref{fig:architecture} and Algorithm~\ref{alg:physic_hm}.

L0: Frozen feature extraction. The framework operates on fixed representations extracted by a suite of frozen state-of-the-art backbones. We utilize V-JEPA2-vitl-fpc64-256 \cite{assran2025vjepa2} to capture long-range motion dynamics from process video , AST-finetuned-audioset-14-14-0.443 \cite{gong21b_interspeech} to provide high-resolution frequency domain acoustic representations , and DINOV3-vith16plus-pretrain-lvd1689m \cite{simeoni2025dinov3} to obtain discriminative semantic features from multi-view post-weld images.

L1: Physic-Hierarchical modulation (PHM). A Mamba-based temporal encoder performs content-aware sequence modeling of high-frequency sensor data with linear complexity. This phase transforms raw sensor signals into governing physical constraints that scale and shift audio-visual features via affine transformations, ensuring that high-dimensional perceptions are conditioned on the underlying production state.

L2: Physic-Hierarchical encoding. Modulated features are integrated into structured semantic spaces to form unified representations. Specifically, a Gated Process Encoder and a Result Encoder generate the process latent  and the result latent , respectively. Within our physical logic,  encapsulates the production input, while  encapsulates the observed outcome.

L3: Anti-Generalization physical decoder. To identify anomalies that violate physical consistency, the model enforces a unidirectional generative mapping . This level incorporates structural safeguards, specifically a Noisy Bottleneck and a Linear Attention Decoder, to prevent the model from learning trivial identity mappings, thereby compelling it to rely on the robust physical laws of normal production rather than brittle feature correlations.

L4: Consistency-Based inference. The model computes an anomaly score  based on the physical consistency between the observed result latent and the prediction generated solely from the process state. By strictly enforcing this physical bottleneck, Physic-HM detects deep-seated industrial defects where the observed effect deviates from the expected outcome predicted by the physical process.

\subsection{Physic-Hierarchical Modulation}
To address the modality drowning effect, we propose the Physic-Hierarchical Modulation (PHM) mechanism. This module treats the production process as a physical governor that conditions the high-dimensional feature space, rather than treating sensors as parallel feature sources. PHM utilizes a Selective State Space Model (Mamba) to capture the long-range temporal dependencies and transient instabilities—such as arc fluctuations or feed speed jitter—inherent in industrial sensor streams. Given the synchronized sensor sequence $X_s \in \mathbb{R}^{T_s \times C_s}$, we employ a Mamba-based temporal encoder to transform the signal into a content-aware latent state $h_s$. This selective mechanism allows the model to dynamically propagate process-critical information based on its relevance to the final weld integrity, effectively encoding a unidirectional physical generative prior from the production process to the resulting effect. Insapired by \cite{perez2018film}, the technical implementation of PHM involves projecting the final hidden state of the Mamba encoder into two distinct affine transformation parameters, $\gamma$ and $\beta \in \mathbb{R}^D$. These parameters act as a physically-grounded constraint that reshapes the manifold of the frozen spatiotemporal features $F_v$ extracted by V-JEPA2 and the acoustic features $F_a$ extracted by AST. The modulated features are formulated as: \begin{equation}
    F_{mod} = F \odot (1 + \gamma) + \beta.
\end{equation}
By utilizing a residual-style scaling factor $(1+\gamma)$, we ensure that the modulation can accurately represent subtle physical deviations—such as a localized current drop—without disrupting the robust semantic information captured by the large-scale frozen backbones. This design enforces a governing relationship where the sensor-driven latent state dictates the activation importance of specific visual and acoustic features. Consequently, the framework remains highly sensitive to process-hidden anomalies that violate physical consistency even when the raw post-weld images appear nominal.

\subsection{Physic-Hierarchical Encoding}
In the physical-hierarchical encoding level, modulated signals are integrated into a unified process representation. For the process modalities, we design a gated process encoder that utilizes a cross-attention mechanism to fuse the spatiotemporal and acoustic streams. In this configuration, the video tokens from V-JEPA2 function as the Query ($Q$), while the audio tokens from AST serve as the key ($K$) and value ($V$). This specific orientation is physically motivated: the model attends to acoustic transients, such as the high-frequency crackling of an unstable arc to modulate the spatial attention of the visual stream. The resulting cross-modal tokens are integrated through a global pooling layer to produce the unified process latent $Z_{p}$, which encapsulates the entirety of the production input within a structured semantic space.

Simultaneously, the framework processes the post-weld images through a result encoder to form the target latent $Z_{r}$, representing the Physical Effect. To handle the complex multi-view nature of industrial inspection, we utilize DINOV3 to extract discriminative features from multiple camera perspectives. These multi-angle representations are aggregated via GatedAngleAggregation, which combines generalized mean (GeM) pooling with a max-pooling operation across the angle dimension. To bridge the heterogeneity between the visual backbone and semantic text constraints, $Z_{r}$ is mapped into a unified hidden dimension ($d=512$), ensuring that the result latent space is aligned for subsequent consistency evaluation. This strategy allows the Result Encoder to remain invariant to camera positioning while being highly sensitive to localized defect signatures that may only appear in a single viewpoint. The hierarchical separation of $Z_{p}$ and $Z_{r}$ ensures that the subsequent mapping stage strictly follows the physical generative logic of the production line. By enforcing this physical bottleneck, the architecture can detect physical consistency breaks during inference, where a seemingly nominal process fails to generate the expected result, or an abnormal process generates a surface that appears visually perfect but violates the learned physical laws of the welding cycle.

\subsection{Anti-Generalization Physical Decoder}
The core of the Physic-HM framework is the mapping from the process latent $Z_{p}$ to the predicted result latent $\hat{Z}_{r}$. In unsupervised anomaly detection, a significant risk is over-generalization, where a powerful decoder learns to reconstruct even anomalous result patterns by memorizing identity mappings. To counter this, we implement two structural safeguards. First, we introduce a Noisy Bottleneck between the encoder and decoder. During training, we apply a stochastic Bernoulli mask and additive Gaussian noise to $Z_{p}$ to form $\hat{Z}_{p}$. This forces the decoder to rely on the robust, high-level physic-logic relationship between the production process and the final product, rather than relying on brittle, high-frequency feature correlations. Second, we replace standard Softmax attention in the decoder with Linear Attention. By utilizing a kernel-based approximation:
\begin{equation}
    \text{Attn}(Q,K,V) = \frac{\phi(Q) \left( \phi(K)^\top V \right)}{\phi(Q) \sum \phi(K)^\top},
\end{equation}
where $\phi(x) = \text{elu}(x) + 1$, the attention mechanism is restricted to a lower-rank representation.
This architectural constraint prevents the decoder from focusing too precisely on local patches, compelling it to reconstruct the result from a global, context-aware perspective of the welding process.

\subsection{Consistency-Based inference}
The model is optimized using a composite loss function designed to enforce physical consistency within the $P \to R$ generative mapping. The primary Reconstruction Loss $\mathcal{L}_{recon}$ integrates cosine distance with a Smooth L1 loss enhanced by Top-K mining. This mechanism focuses optimization on the $k$ dimensions of the latent vector exhibiting the largest absolute errors, thereby increasing sensitivity to localized physical deviations that might otherwise be obscured by global averaging. Furthermore, we introduce a semantic normality constraint $\mathcal{L}_{text}$ to anchor the manifold of normality. To resolve the potential space mismatch between the DINOv3-based latent $Z_{r}$ and the CLIP embedding space, we employ a learnable linear projection layer to align $\hat{Z}_{r}$ with the frozen CLIP text encoder's manifold. This loss ensures the predicted latent remains close to the semantic embedding of the prompt "a normal weld," providing a robust linguistic prior for defect-free production. During inference, the anomaly score $S$ quantifies the violation of the learned physical law by measuring the discrepancy between the observed result and the prediction derived solely from process signals:
\begin{equation}
\begin{split}
S = & 1 - \cos(Z_{r}, \hat{Z}_{r}) \\
& + \eta \cdot \operatorname{mean}(\operatorname{TopK}(|Z_{r} - \hat{Z}_{r}|)),
\end{split}
\end{equation}
where $\eta$ is a weighting hyperparameter. This scoring mechanism identifies physic-logic breaks—scenarios where the observed outcome deviates from the expected effect dictated by the production process. By strictly evaluating the integrity of the $P \to R$ chain, Physic-HM provides a physically-grounded metric that detects deep-seated anomalies invisible to surface-level inspection.

\section{Experiments}

\begin{table*}[t]
    \centering
    \caption{Anomaly detection performance comparison of different methods on Weld-4M dataset. The top 3 best scores for each metric are highlighted as \bfirst{first}, \bsecond{second}, and \bthird{third}.}
    \label{tab:main_results}
    \renewcommand{\arraystretch}{1.05}
    \resizebox{\textwidth}{!}{
    \begin{tabular}{l r rrrrrrrrrrr r r}
        \toprule
        Method & \multicolumn{1}{r}{AUC (\%)} & \multicolumn{1}{r}{\shortstack{Exc.\\Conv.}} & \multicolumn{1}{r}{Undercut} & \multicolumn{1}{r}{\shortstack{Lack of\\Fusion}} & \multicolumn{1}{r}{Poros.} & \multicolumn{1}{r}{Spat.} & \multicolumn{1}{r}{Burn.} & \multicolumn{1}{r}{\shortstack{Poros.\\w/EP}} & \multicolumn{1}{r}{\shortstack{Exc.\\Penet.}} & \multicolumn{1}{r}{\shortstack{Crater\\Cracks}} & \multicolumn{1}{r}{Warp.} & \multicolumn{1}{r}{Over.} & \multicolumn{1}{r}{AP (\%)} & \multicolumn{1}{r}{F1-max (\%)} \\
        \midrule
        LateFusion-Audio & 52.2 & 34.8 & 42.4 & 36.7 & 53.2 & 60.0 & 52.4 & 54.0 & 67.8 & 47.5 & 46.1 & 55.7 & 91.6 & 95.2 \\
        LateFusion-Video & 51.4 & 60.6 & 60.2 & 31.2 & 49.5 & 44.2 & 72.4 & 59.6 & 46.7 & \third{83.4} & 48.6 & 16.2 & 91.7 & 95.3 \\
        LateFusion-Fusion \cite{stemmer2024unsupervised} & 52.8 & 35.4 & 43.1 & 35.4 & 52.6 & 59.7 & 54.7 & 55.3 & 67.9 & 51.6 & 46.2 & 57.0 & 93.0 & 96.4 \\
        BTF \cite{horwitz2023back} & 63.0 & 66.2 & 64.8 & 62.5 & 65.7 & 60.4 & 47.9 & 61.5 & 70.9 & 59.8 & 63.2 & 65.0 & 89.1 & 88.7 \\
        CFM \cite{costanzino2024multimodal} & 67.0 & 68.8 & 67.3 & 65.2 & 70.0 & 63.5 & 58.9 & 66.5 & 71.0 & 64.3 & 65.9 & 67.6 & 95.2 & 94.9 \\
        AST \cite{rudolph2023asymmetric} & 67.6 & 69.8 & 68.2 & 66.5 & 69.1 & 65.4 & 57.5 & 67.1 & 72.2 & 65.0 & 68.3 & 69.7 & 93.7 & 93.5 \\
        M3DM \cite{wang2023multimodal} & 68.2 & 67.6 & 42.4 & 65.6 & 54.5 & \first{95.9} & 61.0 & 74.2 & 59.6 & \first{97.9} & 50.6 & \first{99.6} & 96.8 & 96.4 \\
        3D-ADNAS \cite{long2025revisiting} & 71.9 & 74.2 & 72.8 & 70.2 & 74.0 & 68.5 & 62.9 & 71.5 & 76.0 & 69.3 & 70.9 & 72.6 & \second{98.6} & 97.1 \\
        Reconstruct \cite{guo2023recontrast} & 81.5 & \second{88.7} & \second{87.4} & \third{85.9} & \third{88.5} & 83.8 & \third{76.3} & 85.1 & 90.4 & 82.6 & \third{86.9} & 87.9 & 97.6 & 97.0 \\
        PatchCore \cite{roth2022towards}& 84.2 & 86.7 & 85.9 & 84.5 & 87.3 & 82.6 & 74.8 & 84.1 & 89.2 & 81.3 & 85.8 & 86.9 & 98.1 & \second{97.5} \\
        MVAD \cite{he2024learning} & 85.1 & 87.1 & 86.3 & 85.0 & 87.6 & 83.3 & 75.1 & 84.5 & 89.6 & 82.0 & 85.9 & 86.8 & \third{98.5} & \first{98.0} \\
        RealNet \cite{zhang2024realnet} & 85.3 & \third{87.2} & \third{86.5} & 85.0 & 87.9 & 83.6 & 75.4 & 85.1 & 89.6 & 82.3 & \third{86.9} & 88.2 & 97.0 & 96.5 \\
        MambaAD \cite{he2024mambaad} & 85.3 & \first{89.2} & \first{88.2} & \second{86.4} & \second{89.0} & \third{84.7} & 76.2 & \third{85.9} & \second{93.3} & \second{83.5} & \second{87.5} & 88.9 & 88.5 & 88.0 \\
        RD++ \cite{tien2023revisiting} & 85.4 & 87.1 & 86.3 & 85.0 & 87.6 & 83.3 & 75.3 & 84.5 & 89.4 & 82.0 & 86.5 & 87.9 & 96.5 & 96.1 \\
        UniAD \cite{you2022unified} & 85.6 & \third{87.2} & 86.3 & 84.2 & 88.0 & 82.5 & 73.3 & 83.1 & \third{92.5} & 80.3 & 85.9 & 88.6 & 96.6 & 96.0 \\
        SimpleNet \cite{liu2023simplenet} & 86.9 & 86.1 & 85.4 & 83.9 & 86.9 & 81.3 & 50.5 & 82.7 & 89.0 & 80.5 & 84.8 & 85.9 & 97.0 & 96.5 \\
        ViTAD \cite{zhang2025exploring} & \third{87.1} & 83.2 & 82.3 & 81.0 & 83.6 & 79.3 & 70.3 & 81.5 & 86.5 & 78.0 & 82.9 & 84.8 & 97.2 & 96.6 \\
        Dinomaly \cite{guo2025dinomaly} & \second{88.0} & 78.1 & 81.3 & 85.7 & 81.7 & \second{91.5} & \first{95.7} & \second{88.6} & 91.3 & 81.1 & \first{92.0} & \second{91.2} & \second{98.6} & 97.1 \\
        \textbf{Physic-HM (Ours)} & \first{90.7} & 80.7 & 80.6 & \first{89.3} & \first{93.5} & \first{95.9} & \second{94.9} & \first{97.2} & \first{96.0} & 80.0 & 77.1 & \third{90.8} & \first{99.1} & \third{97.4} \\
        \bottomrule
    \end{tabular}}
    \vspace{-2mm}
\end{table*}

\subsection{Experimental Setup and Dataset}
Existing benchmarks predominantly focus on static post-production images, which fail to capture the underlying physical generative logic and the rich temporal-electrical signals generated during active manufacturing. To evaluate the proposed Physic-HM framework, we utilize the Weld-4M benchmark \cite{stemmer2024unsupervised}, a comprehensive dataset comprising 4,040 samples across 12 categories. Specifically, besides the 'Good' (normal) samples, the dataset encompasses 11 distinct welding defect categories: excessive convexity, undercut, lack of fusion, porosity, spatter, burnthrough, porosity w/EP, excessive penetration, crater cracks, warping, and overlap. For unsupervised training, we utilize 576 defect-free samples. The framework integrates six critical process channels captured in real-time: primary weld current (A), secondary weld voltage (V), wire feed speed (m/min), shielding gas pressure (bar), $CO_2$ flow rate (L/min), and cumulative wire consumption (mm). These sensors operate at an average sampling rate of 9.02 Hz, providing a reliable trace of dynamic process variations. To maintain physical consistency, multimodal synchronization is enforced within a 100ms window using hardware-level absolute timestamps. During data loading, process video is uniformly sampled to 32 frames, while 192 kHz acoustic emissions and sensor time-series are resampled to 256 temporal bins to maintain physical consistency within a 100ms synchronization window. Post-weld inspection is conducted via five independent camera angles captured after welding completion. Performance was quantified using image-level AUROC, average precision (I-AP), and best F1 score (I-F1-max), with each metric being the average of 5 replicate experiments to ensure rigorous and comprehensive evaluation.

\subsection{Baselines}

To rigorously validate the effectiveness of our physical-logic approach, we compare Physic-HM against 18 baseline methods spanning diverse research paradigms. Since no existing unsupervised anomaly detection (UAD) framework natively supports the four distinct industrial modalities (video, audio, sensors, and post-weld images) provided by Weld-4M, we selected representative baselines that natively support the result modality (post-weld images) at least, as image-level inspection remains the industry standard for identifying manifest defects.
(1) Unimodal Baselines: We evaluate single-modality performance using Dinomaly \cite{guo2025dinomaly} for multi-class visual detection and AST \cite{rudolph2023asymmetric} for acoustic signal analysis. (2) Simple Fusion Strategies: We implement score-level LateFusion (across Audio, Video, and Fusion) following \cite{stemmer2024unsupervised} to provide a baseline for symmetric multimodal integration. (3) Adapted Multimodal Detectors: To adapt SOTA RGB-D detectors such as M3DM \cite{wang2023multimodal}, 3D-ADNAS \cite{long2025revisiting} and BTF \cite{horwitz2023back}, we transformed 1D sensor time-series into pseudo-point cloud tensors to satisfy their geometric architectural requirements, details are provided in \ref{sec:implementation}. (4) Reconstruction-based and Mapping Models: The comparison includes CFM \cite{costanzino2024multimodal}, which performs cross-modal feature mapping, alongside advanced reconstruction frameworks such as RealNet \cite{zhang2024realnet}, Reconstruct \cite{guo2023recontrast}, ViTAD \cite{zhang2025exploring}, and MambaAD \cite{he2024mambaad}. (5) Embedding and Distributional Models: We further include PatchCore \cite{roth2022towards}, MVAD \cite{he2024learning}, RD++ \cite{tien2023revisiting}, UniAD \cite{you2022unified}, and SimpleNet \cite{liu2023simplenet} to represent the diversity of feature-embedding and normalizing flow paradigms.

\subsection{Implementation Details} \label{sec:implementation}
Physic-HM and all baselines are trained and evaluated on a single NVIDIA RTX 4090D GPU and we set a fixed random seed of 42 for all trials. All trainable components are optimized using the AdamW optimizer with a learning rate of $1 \times 10^{-4}$ and a batch size of 16 for 300 epochs. We employ a cosine annealing scheduler featuring no warmup to stabilize the initial physical generative mapping. Following the unsupervised paradigm \cite{stemmer2024unsupervised}, the training set comprises 576 "Good" samples and the validation set includes 122 "Good" and 1,610 "Defective" samples. We then report the final performance on the test set, which consists of 121 "Good" and 1,611 "Defective" samples.

To ensure a fair comparison with RGBD baselines like M3DM and BTF, we map 1D sensor time-series into pseudo-point cloud tensors, and treat the sensor modalities as a structured geometric surface. Specifically, for a synchronized sequence of $C$ sensor channels across $T$ time steps, we construct a meshgrid where the $X$ and $Y$ coordinates represent the normalized time indices and channel indices, respectively. The normalized magnitude of each sensor signal is assigned as the depth value ($Z$). The resulting $T \times C$ representation is then upsampled via bilinear interpolation to match the backbone's input resolution (e.g., $224 \times 224$). This mapping preserves the unique temporal-electrical signatures of individual sensors while satisfying the geometric architectural requirements of the baselines, ensuring that the performance gains of Physic-HM stem from its physical-hierarchical logic rather than an informational disadvantage of the competing models.

\subsection{Main Results} 
The quantitative comparison results summarized in Table~\ref{tab:main_results} reveal that Physic-HM achieves a significant performance leap, reaching a SOTA I-AUROC of 90.7\% on the test set. Our method outperforms the strongest baseline, Dinomaly, by a substantial margin of 2.7\%. Deep insights into the categorical breakdown show that while visual-only models like Dinomaly perform adequately on surface-level defects, their sensitivity drops drastically when encountering process-hidden defects such as Lack of Fusion and Porosity. By contrast, Physic-HM maintains high sensitivity by capturing anomalies in the process-result physical chain that are otherwise invisible to surface inspection. This superiority is attributed to the physical inductive bias, which forces the model to learn the physical law of normal welding rather than simple statistical correlations. Furthermore, our method achieves an I-AP of 99.1\% and an I-F1-max of 97.4\%, suggesting that the hierarchical separation of process and effect provides a more resilient manifold for industrial anomaly detection.

\begin{table}[t]
    \centering
    \caption{Unified modality and component ablation. AUC-gain is computed relative to the \textit{image+video+audio+sensor} setting.}
    \label{tab:ablations}
    {
    \resizebox{\columnwidth}{!}{
    \begin{tabular}{lrrrr}
        \toprule
        Setting & \multicolumn{1}{r}{I-AUROC (\%)} & \multicolumn{1}{r}{AUC-gain} & \multicolumn{1}{r}{AP (\%)} & \multicolumn{1}{r}{F1-max (\%)} \\
        \midrule
        Reverse Mapping ($R \to P$) & 82.9 & $-7.8$ & 98.3 & 96.6 \\
        Plain Decoder & 83.2 & $-7.5$ & 98.2 & 96.4 \\
        No CLIP $L_{\text{text}}$ & 83.4 & $-7.3$ & 98.2 & 96.4 \\
        Bidirectional ($P \leftrightarrow R$) & 87.2 & $-3.5$ & 98.5 & 96.9 \\
        Symmetric Fusion (Sensor) & 89.3 & $-1.4$ & 98.8 & 96.9 \\
        \midrule
        Image & 87.0 & $-3.7$ & 98.6 & 97.4 \\
        Image + Video & 87.5 & $-3.2$ & 98.4 & 96.8 \\
        Image + Video + Audio & 90.0 & $-0.7$ & 98.7 & 96.9 \\
        Image + Video + Audio + Sensor & 90.7 & $0.0$ & 99.1 & 97.4 \\
        \bottomrule
    \end{tabular}}}
    \vspace{-2mm}
\end{table}

\subsection{Ablation Experiments}

Systematic ablations were conducted to validate our architectural design choices, with results consolidated in Table~\ref{tab:ablations}. We first investigate the contribution of different modalities: starting from a baseline I-AUROC of 87.0\% using only images, the steady increase to 87.5\% (Image+Video) and 90.0\% (Image+Video+Audio) confirms that each additional modality injects complementary physical cues into the representation. Compared to a symmetric integration baseline that simply concatenates sensor features (Symmetric Fusion, 89.3\%), the inclusion of our proposed PHM-mediated sensor modulation delivers the final performance gain, reaching the peak I-AUROC of 90.7\%. Regarding architectural variants, reversing the flow to predict the process from the result (reverse mapping $R \rightarrow P$) leads to a significant performance drop of 7.8\%, which empirically validates that the unidirectional $P \rightarrow R$ mapping is more consistent with the physical generative logic of industrial production than its inverse. Furthermore, removing the anti-generalization safeguards (the noisy bottleneck and linear attention) results in a degradation of 7.5\%. These components are critical for preventing trivial identity mappings by compelling the model to rely on lower-rank representations and global process context rather than brittle local correlations. Finally, the exclusion of the semantic CLIP alignment leads to a 7.3\% drop, confirming its necessity as a linguistic anchor for defining the manifold of normality.

\begin{table}
    \centering
    \caption{Inference efficiency and memory footprint comparison (all methods use 4 modalities).}
    \label{tab:efficiency}
    \resizebox{0.9\columnwidth}{!}{
    \begin{tabular}{lrrr} 
        \toprule
        Method & I-AUROC & Latency (ms) & FPS \\ 
        \midrule
        M3DM             & 68.2 & 1567.0 & 0.6 \\ 
        PatchCore        & 84.2 &  520.0 & 1.9 \\
        \textbf{Physic-HM (Ours)} & \textbf{90.7} & \textbf{268.0} & \textbf{3.7} \\
        \bottomrule
    \end{tabular}}
\end{table}

\subsection{Efficiency and Inference Robustness} 
\begin{figure}[htb]
	\centering
	\includegraphics[width=\linewidth]{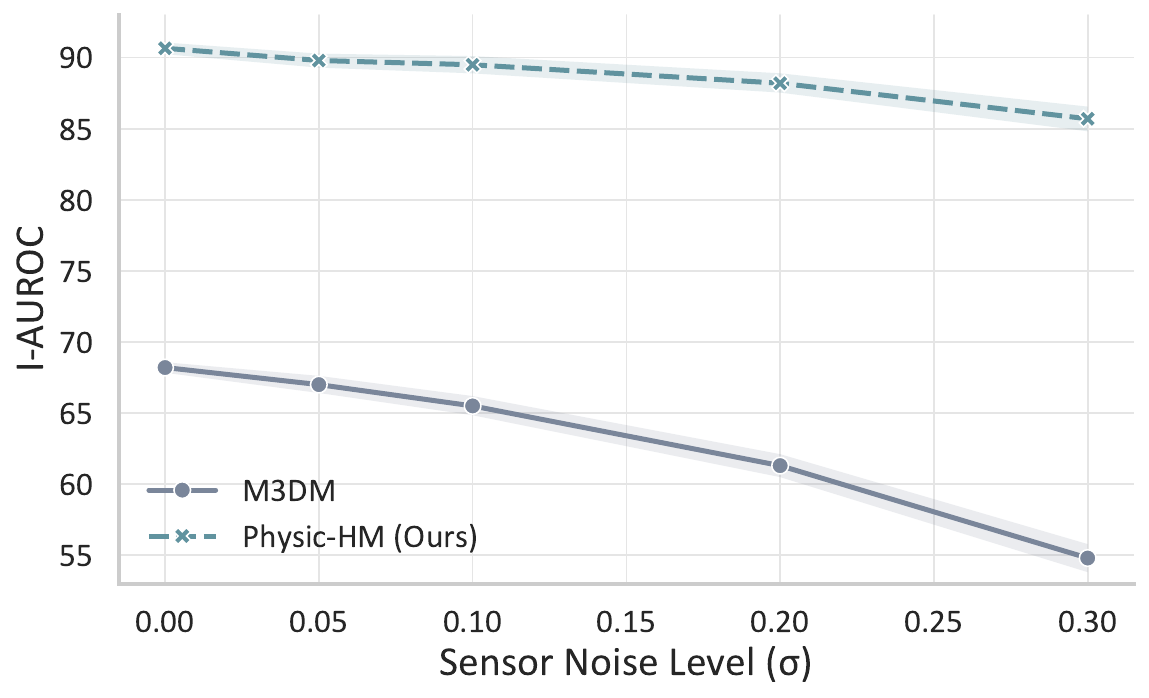}
    \vspace{-5mm}
    \caption{Robustness analysis under environmental noise.}
    \label{fig:robustness}
\end{figure}

Industrial deployment feasibility is assessed by analyzing the trade-off between detection accuracy and computational overhead. As detailed in Table~\ref{tab:efficiency}, despite a larger memory footprint, Physic-HM achieves an inference speed of 3.7 FPS, making it nearly 6 times faster than the M3DM baseline (0.6 FPS). This significant efficiency gain is attributed to the replacement of computationally expensive memory bank retrievals with our streamlined Physic-Hierarchical mapping decoder. To further assess the robustness of the proposed architecture, we evaluate detection performance under varying levels of environmental sensor noise. As illustrated in Figure~\ref{fig:robustness}, Physic-HM exhibits superior resilience; even at a high sensor noise level ($\sigma=0.3$), it maintains a robust I-AUROC of 85.7\%, whereas M3DM's performance collapses to 54.8\%. These results validate that explicitly modeling the physical generative logic allows the system to leverage the Process $\rightarrow$ Result dependency to compensate for modality-specific noise or occlusions.

\begin{figure}[htb]\centering\includegraphics[width=\linewidth]{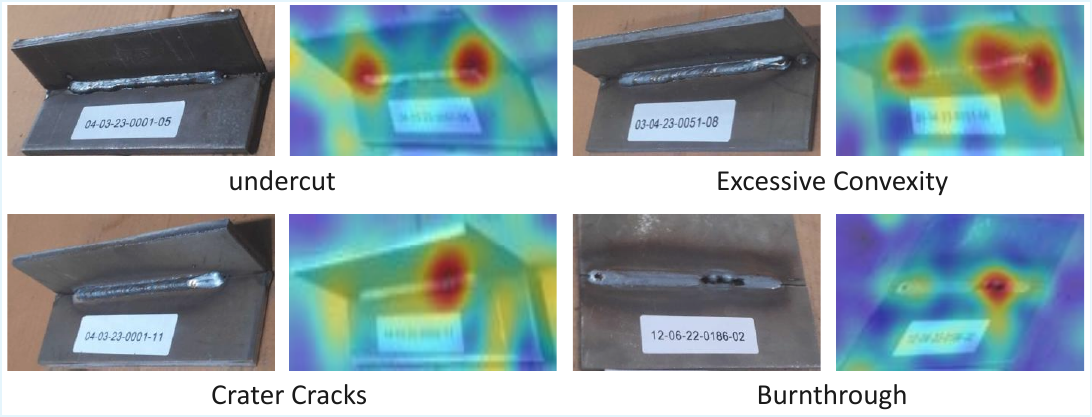}
\caption{Qualitative results comparing sampled image and our predicted anomaly map.}
\label{Qualitative}
\end{figure}

\subsection{Case Study}
Figure~\ref{Qualitative} demonstrates the framework's capability to accurately localize various explicit surface anomalies within the Weld-4M benchmark, including undercut, excessive convexity, crater cracks, and burnthrough. The generated anomaly maps exhibit high response intensity specifically in regions where the physical consistency of the $P \to R$ mapping is violated. For details of its generation, see in Supplementary Material~\ref{appendix:heatmap}. This confirms that Physic-HM does not merely perform a symmetric aggregation of multimodal features but actively evaluates the physical integrity of the entire production sequence. Such a dual capability—detecting deep-seated process anomalies that evade traditional visual inspection while maintaining high precision on manifest surface defects—establishes Physic-HM as a robust solution for high-reliability industrial quality assurance.

\section{Conclusions}
In this paper, we propose Physic-HM, a multimodal UAD framework that explicitly incorporates physical generative priors to model the unidirectional Process $\rightarrow$ Result dependency in industrial manufacturing. By transitioning from traditional flat fusion to a Physic-Hierarchical architecture, we address process-logic blindness and effectively bridge the heterogeneity gap between low-dimensional sensor signals and high-dimensional visual streams. Our core innovations—Sensor-Guided PHM modulation and an anti-generalization decoder—leverage physically-grounded inductive biases to identify anomalies that violate production consistency while suppressing the learning of trivial identity mappings. Extensive evaluation on the Weld-4M benchmark demonstrates that Physic-HM achieves a SOTA I-AUROC of 90.7\%, exhibiting superior sensitivity to deep-seated process defects that evade surface-level inspection while maintaining high inference efficiency and Robustness.

\bibliographystyle{named}
\bibliography{ijcai26}

\begin{thebibliography}{}

\bibitem[\protect\citeauthoryear{Assran \bgroup \em et al.\egroup }{2025}]{assran2025vjepa2}
Mahmoud Assran, Adrien Bardes, David Fan, Quentin Garrido, Russell Howes, Mojtaba Komeili, Matthew Muckley, Ammar Rizvi, Claire Roberts, Koustuv Sinha, Artem Zholus, Sergio Arnaud, Abha Gejji, Ada Martin, Francois Robert~Hogan, Daniel Dugas, Piotr Bojanowski, Vasil Khalidov, Patrick Labatut, Francisco Massa, Marc Szafraniec, Kapil Krishnakumar, Yong Li, Xiaodong Ma, Sarath Chandar, Franziska Meier, Yann LeCun, Michael Rabbat, and Nicolas Ballas.
\newblock V-jepa~2: Self-supervised video models enable understanding, prediction and planning.
\newblock Technical report, FAIR at Meta, 2025.

\bibitem[\protect\citeauthoryear{Barua \bgroup \em et al.\egroup }{2023}]{barua2023systematic}
Arnab Barua, Mobyen~Uddin Ahmed, and Shahina Begum.
\newblock A systematic literature review on multimodal machine learning: Applications, challenges, gaps and future directions.
\newblock {\em Ieee access}, 11:14804--14831, 2023.

\bibitem[\protect\citeauthoryear{Cheng \bgroup \em et al.\egroup }{2023}]{cheng2023intelligent}
Xiang Cheng, Haotian Zhang, Jianan Zhang, Shijian Gao, Sijiang Li, Ziwei Huang, Lu~Bai, Zonghui Yang, Xinhu Zheng, and Liuqing Yang.
\newblock Intelligent multi-modal sensing-communication integration: Synesthesia of machines.
\newblock {\em IEEE Communications Surveys \& Tutorials}, 26(1):258--301, 2023.

\bibitem[\protect\citeauthoryear{Cheng \bgroup \em et al.\egroup }{2025}]{cheng2025mc3d}
Jiayi Cheng, Can Gao, Jie Zhou, Jiajun Wen, Tao Dai, and Jinbao Wang.
\newblock Mc3d-ad: A unified geometry-aware reconstruction model for multi-category 3d anomaly detection.
\newblock {\em arXiv preprint arXiv:2505.01969}, 2025.

\bibitem[\protect\citeauthoryear{Costanzino \bgroup \em et al.\egroup }{2024}]{costanzino2024multimodal}
Alex Costanzino, Pierluigi~Zama Ramirez, Giuseppe Lisanti, and Luigi Di~Stefano.
\newblock Multimodal industrial anomaly detection by crossmodal feature mapping.
\newblock In {\em Proceedings of the IEEE/CVF Conference on Computer Vision and Pattern Recognition}, pages 17234--17243, 2024.

\bibitem[\protect\citeauthoryear{Costanzino \bgroup \em et al.\egroup }{2025}]{costanzino2025sim3d}
Alex Costanzino, Pierluigi~Zama Ramirez, Luigi Lella, Matteo Ragaglia, Alessandro Oliva, Giuseppe Lisanti, and Luigi Di~Stefano.
\newblock Sim3d: Single-instance multiview multimodal and multisetup 3d anomaly detection benchmark.
\newblock {\em arXiv preprint arXiv:2506.21549}, 2025.

\bibitem[\protect\citeauthoryear{Gao \bgroup \em et al.\egroup }{2024}]{gao2024embracing}
Zixian Gao, Xun Jiang, Xing Xu, Fumin Shen, Yujie Li, and Heng~Tao Shen.
\newblock Embracing unimodal aleatoric uncertainty for robust multimodal fusion.
\newblock In {\em Proceedings of the IEEE/CVF conference on computer vision and pattern recognition}, pages 26876--26885, 2024.

\bibitem[\protect\citeauthoryear{Gong \bgroup \em et al.\egroup }{2021}]{gong21b_interspeech}
Yuan Gong, Yu-An Chung, and James Glass.
\newblock {AST: Audio Spectrogram Transformer}.
\newblock In {\em Proc. Interspeech 2021}, pages 571--575, 2021.

\bibitem[\protect\citeauthoryear{Gu and Dao}{2024}]{gu2024mamba}
Albert Gu and Tri Dao.
\newblock Mamba: Linear-time sequence modeling with selective state spaces.
\newblock In {\em First conference on language modeling}, 2024.

\bibitem[\protect\citeauthoryear{Gu \bgroup \em et al.\egroup }{2024}]{gu2024rethinking}
Zhihao Gu, Jiangning Zhang, Liang Liu, Xu~Chen, Jinlong Peng, Zhenye Gan, Guannan Jiang, Annan Shu, Yabiao Wang, and Lizhuang Ma.
\newblock Rethinking reverse distillation for multi-modal anomaly detection.
\newblock In {\em Proceedings of the AAAI Conference on Artificial Intelligence}, volume~38, pages 8445--8453, 2024.

\bibitem[\protect\citeauthoryear{Guo \bgroup \em et al.\egroup }{2023}]{guo2023recontrast}
Jia Guo, Shuai Lu, Lize Jia, Weihang Zhang, and Huiqi Li.
\newblock Recontrast: Domain-specific anomaly detection via contrastive reconstruction.
\newblock {\em Advances in Neural Information Processing Systems}, 36:10721--10740, 2023.

\bibitem[\protect\citeauthoryear{Guo \bgroup \em et al.\egroup }{2025}]{guo2025dinomaly}
Jia Guo, Shuai Lu, Weihang Zhang, Fang Chen, Huiqi Li, and Hongen Liao.
\newblock Dinomaly: The less is more philosophy in multi-class unsupervised anomaly detection.
\newblock In {\em Proceedings of the Computer Vision and Pattern Recognition Conference}, pages 20405--20415, 2025.

\bibitem[\protect\citeauthoryear{He \bgroup \em et al.\egroup }{2024a}]{he2024mambaad}
Haoyang He, Yuhu Bai, Jiangning Zhang, Qingdong He, Hongxu Chen, Zhenye Gan, Chengjie Wang, Xiangtai Li, Guanzhong Tian, and Lei Xie.
\newblock Mambaad: Exploring state space models for multi-class unsupervised anomaly detection.
\newblock {\em Advances in Neural Information Processing Systems}, 37:71162--71187, 2024.

\bibitem[\protect\citeauthoryear{He \bgroup \em et al.\egroup }{2024b}]{he2024learning}
Haoyang He, Jiangning Zhang, Guanzhong Tian, Chengjie Wang, and Lei Xie.
\newblock Learning multi-view anomaly detection.
\newblock {\em arXiv preprint arXiv:2407.11935}, 1(2):3, 2024.

\bibitem[\protect\citeauthoryear{Hong \bgroup \em et al.\egroup }{2024}]{hong2024af}
Yuxiang Hong, Xingxing He, Jing Xu, Ruiling Yuan, Kai Lin, Baohua Chang, and Dong Du.
\newblock Af-fttsnet: An end-to-end two-stream convolutional neural network for online quality monitoring of robotic welding.
\newblock {\em Journal of Manufacturing Systems}, 74:422--434, 2024.

\bibitem[\protect\citeauthoryear{Horwitz and Hoshen}{2023}]{horwitz2023back}
Eliahu Horwitz and Yedid Hoshen.
\newblock Back to the feature: classical 3d features are (almost) all you need for 3d anomaly detection.
\newblock In {\em Proceedings of the IEEE/CVF Conference on Computer Vision and Pattern Recognition}, pages 2968--2977, 2023.

\bibitem[\protect\citeauthoryear{Kim \bgroup \em et al.\egroup }{2024}]{kim2024rethinking}
Sunwoo Kim, Soo~Yong Lee, Fanchen Bu, Shinhwan Kang, Kyungho Kim, Jaemin Yoo, and Kijung Shin.
\newblock Rethinking reconstruction-based graph-level anomaly detection: limitations and a simple remedy.
\newblock {\em Advances in Neural Information Processing Systems}, 37:95931--95962, 2024.

\bibitem[\protect\citeauthoryear{Li \bgroup \em et al.\egroup }{2025}]{li2025multi}
Wenqiao Li, Bozhong Zheng, Xiaohao Xu, Jinye Gan, Fading Lu, Xiang Li, Na~Ni, Zheng Tian, Xiaonan Huang, Shenghua Gao, et~al.
\newblock Multi-sensor object anomaly detection: Unifying appearance, geometry, and internal properties.
\newblock In {\em Proceedings of the Computer Vision and Pattern Recognition Conference}, pages 9984--9993, 2025.

\bibitem[\protect\citeauthoryear{Liang \bgroup \em et al.\egroup }{2023}]{liang2023omni}
Yufei Liang, Jiangning Zhang, Shiwei Zhao, Runze Wu, Yong Liu, and Shuwen Pan.
\newblock Omni-frequency channel-selection representations for unsupervised anomaly detection.
\newblock {\em IEEE Transactions on Image Processing}, 32:4327--4340, 2023.

\bibitem[\protect\citeauthoryear{Lin \bgroup \em et al.\egroup }{2025}]{lin2025survey}
Yuxuan Lin, Yang Chang, Xuan Tong, Jiawen Yu, Antonio Liotta, Guofan Huang, Wei Song, Deyu Zeng, Zongze Wu, Yan Wang, et~al.
\newblock A survey on rgb, 3d, and multimodal approaches for unsupervised industrial image anomaly detection.
\newblock {\em Information Fusion}, page 103139, 2025.

\bibitem[\protect\citeauthoryear{Liu \bgroup \em et al.\egroup }{2023}]{liu2023simplenet}
Zhikang Liu, Yiming Zhou, Yuansheng Xu, and Zilei Wang.
\newblock Simplenet: A simple network for image anomaly detection and localization.
\newblock In {\em Proceedings of the IEEE/CVF conference on computer vision and pattern recognition}, pages 20402--20411, 2023.

\bibitem[\protect\citeauthoryear{Liu \bgroup \em et al.\egroup }{2024a}]{liu2024deep}
Jiaqi Liu, Guoyang Xie, Jinbao Wang, Shangnian Li, Chengjie Wang, Feng Zheng, and Yaochu Jin.
\newblock Deep industrial image anomaly detection: A survey.
\newblock {\em Machine Intelligence Research}, 21(1):104--135, 2024.

\bibitem[\protect\citeauthoryear{Liu \bgroup \em et al.\egroup }{2024b}]{liu2024arc}
Yixin Liu, Shiyuan Li, Yu~Zheng, Qingfeng Chen, Chengqi Zhang, and Shirui Pan.
\newblock Arc: A generalist graph anomaly detector with in-context learning.
\newblock {\em Advances in Neural Information Processing Systems}, 37:50772--50804, 2024.

\bibitem[\protect\citeauthoryear{Long \bgroup \em et al.\egroup }{2025}]{long2025revisiting}
Kaifang Long, Guoyang Xie, Lianbo Ma, Jiaqi Liu, and Zhichao Lu.
\newblock Revisiting multimodal fusion for 3d anomaly detection from an architectural perspective.
\newblock In {\em Proceedings of the AAAI Conference on Artificial Intelligence}, volume~39, pages 12273--12281, 2025.

\bibitem[\protect\citeauthoryear{Niu \bgroup \em et al.\egroup }{2025}]{niu2024zero}
Chaoxi Niu, Hezhe Qiao, Changlu Chen, Ling Chen, and Guansong Pang.
\newblock Zero-shot generalist graph anomaly detection with unified neighborhood prompts.
\newblock In {\em Proceedings of the Thirty-Fourth International Joint Conference on Artificial Intelligence}, IJCAI '25, 2025.

\bibitem[\protect\citeauthoryear{Pemula \bgroup \em et al.\egroup }{2025}]{pemula2025robust}
Latha Pemula, Dongqing Zhang, and Onkar Dabeer.
\newblock Robust ad: A real world benchmark dataset for robustness in industrial anomaly detection.
\newblock In {\em Proceedings of the Computer Vision and Pattern Recognition Conference}, pages 4047--4057, 2025.

\bibitem[\protect\citeauthoryear{Perez \bgroup \em et al.\egroup }{2018}]{perez2018film}
Ethan Perez, Florian Strub, Harm De~Vries, Vincent Dumoulin, and Aaron Courville.
\newblock Film: Visual reasoning with a general conditioning layer.
\newblock In {\em Proceedings of the AAAI conference on artificial intelligence}, volume~32, 2018.

\bibitem[\protect\citeauthoryear{Roth \bgroup \em et al.\egroup }{2022}]{roth2022towards}
Karsten Roth, Latha Pemula, Joaquin Zepeda, Bernhard Sch{\"o}lkopf, Thomas Brox, and Peter Gehler.
\newblock Towards total recall in industrial anomaly detection.
\newblock In {\em Proceedings of the IEEE/CVF conference on computer vision and pattern recognition}, pages 14318--14328, 2022.

\bibitem[\protect\citeauthoryear{Rudolph \bgroup \em et al.\egroup }{2023}]{rudolph2023asymmetric}
Marco Rudolph, Tom Wehrbein, Bodo Rosenhahn, and Bastian Wandt.
\newblock Asymmetric student-teacher networks for industrial anomaly detection.
\newblock In {\em Proceedings of the IEEE/CVF winter conference on applications of computer vision}, pages 2592--2602, 2023.

\bibitem[\protect\citeauthoryear{Sim{\'e}oni \bgroup \em et al.\egroup }{2025}]{simeoni2025dinov3}
Oriane Sim{\'e}oni, Huy~V. Vo, Maximilian Seitzer, Federico Baldassarre, Maxime Oquab, Cijo Jose, Vasil Khalidov, Marc Szafraniec, Seungeun Yi, Micha{\"e}l Ramamonjisoa, Francisco Massa, Daniel Haziza, Luca Wehrstedt, Jianyuan Wang, Timoth{\'e}e Darcet, Th{\'e}o Moutakanni, Leonel Sentana, Claire Roberts, Andrea Vedaldi, Jamie Tolan, John Brandt, Camille Couprie, Julien Mairal, Herv{\'e} J{\'e}gou, Patrick Labatut, and Piotr Bojanowski.
\newblock {DINOv3}, 2025.

\bibitem[\protect\citeauthoryear{Stemmer \bgroup \em et al.\egroup }{2024}]{stemmer2024unsupervised}
Georg Stemmer, Jose~A Lopez, Juan~A Ontiveros, Arvind Raju, Tara Thimmanaik, and Sovan Biswas.
\newblock Unsupervised welding defect detection using audio and video.
\newblock {\em arXiv preprint arXiv:2409.02290}, 2024.

\bibitem[\protect\citeauthoryear{Tien \bgroup \em et al.\egroup }{2023}]{tien2023revisiting}
Tran~Dinh Tien, Anh~Tuan Nguyen, Nguyen~Hoang Tran, Ta~Duc Huy, Soan Duong, Chanh D~Tr Nguyen, and Steven~QH Truong.
\newblock Revisiting reverse distillation for anomaly detection.
\newblock In {\em Proceedings of the IEEE/CVF conference on computer vision and pattern recognition}, pages 24511--24520, 2023.

\bibitem[\protect\citeauthoryear{Wang \bgroup \em et al.\egroup }{2023}]{wang2023multimodal}
Yue Wang, Jinlong Peng, Jiangning Zhang, Ran Yi, Yabiao Wang, and Chengjie Wang.
\newblock Multimodal industrial anomaly detection via hybrid fusion.
\newblock In {\em Proceedings of the IEEE/CVF conference on computer vision and pattern recognition}, pages 8032--8041, 2023.

\bibitem[\protect\citeauthoryear{Wang \bgroup \em et al.\egroup }{2025}]{wang2025m3dm}
Chengjie Wang, Haokun Zhu, Jinlong Peng, Yue Wang, Ran Yi, Yunsheng Wu, Lizhuang Ma, and Jiangning Zhang.
\newblock M3dm-nr: Rgb-3d noisy-resistant industrial anomaly detection via multimodal denoising.
\newblock {\em IEEE Transactions on Pattern Analysis and Machine Intelligence}, 2025.

\bibitem[\protect\citeauthoryear{Wu \bgroup \em et al.\egroup }{2024}]{wu2024cross}
Gaochang Wu, Yapeng Zhang, Lan Deng, Jingxin Zhang, and Tianyou Chai.
\newblock Cross-modal learning for anomaly detection in complex industrial process: Methodology and benchmark.
\newblock {\em IEEE Transactions on Circuits and Systems for Video Technology}, 2024.

\bibitem[\protect\citeauthoryear{You \bgroup \em et al.\egroup }{2022}]{you2022unified}
Zhiyuan You, Lei Cui, Yujun Shen, Kai Yang, Xin Lu, Yu~Zheng, and Xinyi Le.
\newblock A unified model for multi-class anomaly detection.
\newblock {\em Advances in Neural Information Processing Systems}, 35:4571--4584, 2022.

\bibitem[\protect\citeauthoryear{Zhang \bgroup \em et al.\egroup }{2024a}]{zhang2024contextual}
Jie Zhang, Masanori Suganuma, and Takayuki Okatani.
\newblock Contextual affinity distillation for image anomaly detection.
\newblock In {\em Proceedings of the IEEE/CVF Winter Conference on Applications of Computer Vision}, pages 149--158, 2024.

\bibitem[\protect\citeauthoryear{Zhang \bgroup \em et al.\egroup }{2024b}]{zhang2024realnet}
Ximiao Zhang, Min Xu, and Xiuzhuang Zhou.
\newblock Realnet: A feature selection network with realistic synthetic anomaly for anomaly detection.
\newblock In {\em Proceedings of the IEEE/CVF conference on computer vision and pattern recognition}, pages 16699--16708, 2024.

\bibitem[\protect\citeauthoryear{Zhang \bgroup \em et al.\egroup }{2025}]{zhang2025exploring}
Jiangning Zhang, Xuhai Chen, Yabiao Wang, Chengjie Wang, Yong Liu, Xiangtai Li, Ming-Hsuan Yang, and Dacheng Tao.
\newblock Exploring plain vit features for multi-class unsupervised visual anomaly detection.
\newblock {\em Computer Vision and Image Understanding}, 253:104308, 2025.

\bibitem[\protect\citeauthoryear{Zhou \bgroup \em et al.\egroup }{2023}]{zhou2023anomalyclip}
Qihang Zhou, Guansong Pang, Yu~Tian, Shibo He, and Jiming Chen.
\newblock Anomalyclip: Object-agnostic prompt learning for zero-shot anomaly detection.
\newblock {\em arXiv preprint arXiv:2310.18961}, 2023.

\bibitem[\protect\citeauthoryear{Zhu and Pang}{2024}]{zhu2024toward}
Jiawen Zhu and Guansong Pang.
\newblock Toward generalist anomaly detection via in-context residual learning with few-shot sample prompts.
\newblock In {\em Proceedings of the IEEE/CVF conference on computer vision and pattern recognition}, pages 17826--17836, 2024.

\end{thebibliography}

\appendix
\section{Anomaly Heatmap Generation} \label{appendix:heatmap}
To provide pixel-level localization of industrial defects, Physic-HM employs a gradient-based saliency mapping technique. Following Dinomaly \cite{guo2025dinomaly}, Our framework derives the anomaly heatmap by backpropagating the consistency-based anomaly score $S$ directly to the feature tokens of the frozen DINOv3 backbone. Specifically, let $F_i$ be the set of visual tokens; the pixel-level response is defined by the gradient $\nabla_{F_i} S$, where $S = 1 - \cos(Z_r, \hat{Z}_r)$ represents the violation of the learned physical mapping. Since the reference latent $\hat{Z}_r$ is dynamically generated from process modalities (sensors, audio, and real-time video), the resulting gradients capture the precise spatial regions where the observed post-weld surface deviates from the physically expected outcome dictated by the production process. These raw saliency maps undergo spatial reshaping, bilinear interpolation to the original resolution, and Gaussian smoothing to ensure visual coherence. This mechanism ensures that the localized "hotspots" are not merely visual outliers but are physically-grounded evidence of process-to-result consistency breaks.

\section{Declaration of Generative AI Usage}
During the preparation of this work, we used generative ai exclusively for language editing, grammatical refinement, and stylistic polishing to improve the manuscript's readability. We declare that no generative AI was used in the core scientific contributions of this research. Specifically, the conceptualization of the Physic-HM framework, the mathematical formulation of the Physic-Hierarchical Modulation (PHM) mechanism, the experimental design and execution on the Weld-4M benchmark, and the interpretation of the results were authored entirely by the human authors. The authors assume full responsibility for the content of the manuscript, including its technical correctness, factual accuracy, and original integrity, and have verified the text to ensure the absence of plagiarism.

\end{document}